# A distributed Approach for Access and Visibility Task with a Manikin and a Robot in a Virtual Reality Environment

P. Chedmail, D. Chablat, and Ch. Le Roy

*Abstract*— **This paper presents a new method, based on a multi-agent system and on a digital mock-up technology, to assess an efficient path planner for a manikin or a robot for access and visibility task taking into account ergonomic constraints or joint and mechanical limits. In order to solve this problem, the human operator is integrated in the process optimization to contribute to a global perception of the environment. This operator cooperates, in real-time, with several automatic local elementary agents. The result of this work validates solutions through the digital mock-up; it can be applied to simulate maintenability and mountability tasks.**

## I. INTRODUCTION

In an industrial environment, the access to a sharable and global view of the enterprise project, product, and/or service appears to be a key factor of success. It improves the triptych delay-quality-cost but also the communication between the different partners and their implication in the project. For these reasons, the digital mock-up (DMU) and its functions are deeply investigated by industrials. Based on computer technology and virtual reality, the DMU consists in a platform of visualization and simulation that can cover different processes and areas during the product lifecycle such as product design, industrialization, production, maintenance, recycling and/or customer support (Figure 1).

The digital model enables the earlier identification of possible issues and a better understanding of the processes even, and maybe above all, for actors who are not specialists. Thus, a digital model allows deciding before expensive physical prototypes have been built. Even if evident progresses were noticed and applied in the domain of DMUs, significant progresses are still awaited for a placement in an industrial context. As a matter of fact, the digital model offers a way to explore areas such as maintenance or ergonomics of the product that were traditionally ignored at the beginning phases of a project; new processes must consequently be developed.

Through the integration of a manikin or a robot in a virtual environment, the suitability of a product, its shape and functions can be assessed. In the same time, it becomes possible to settle the process for assembling with a robot the different components of the product. Moreover, when simulating a task that should be performed by an operator with a virtual manikin model, feasibility, access and visibility can be checked. The conditions of the performances in terms of efforts, constraints and comfort can also be analyzed. Modifications on the process, on the product or on the task itself may follow but also a better and earlier training of the operators to enhance their performances in the real environment. Moreover, such a use of the DMU leads to a better conformance to health and safety standards, to a maximization of human comfort and safety and an optimization of the robot abilities.

With virtual reality tools such as 3D manipulators (Figure 2), it is possible to manipulate the object as easily as in a real to manipulate the object as easily as in a real environment. Some drawbacks are the difficulty to manipulate the object with as ease as in a real environment, due to the lack of kinematics constraints and the automatic collision avoidance. As a matter of fact, interference detection between parts is often displayed through color changes of parts in collision but collision is not avoided.

Another approach consists in integrating automatic functionality into the virtual environment in order to ease the user's task. Many research topics in the framework of robotics dealing with the definition of collision-free trajectories for solid objects are also valid in the DMU. Some methodologies need a global perception of the environment, like (i) visibility graphs proposed by Lozano-Pérez and Wesley [1], (ii) geodesic graphs proposed by Tournassoud [2], or (iii) Voronoï's diagrams [3]. However, these techniques are very CPU consuming but lead to a solution if it exists. Some other methodologies consider the moves of the object only in its close or local environment. The success of these methods is not guaranteed due to the existence of local minima. A specific method was proposed by Khatib [4] and enhanced by Barraquand and Latombe [5]. In this method, Khatib's potentials method is coupled with an optimization method that minimizes the distance to the target and avoids collisions. All these techniques are limited, either by the computation cost, or the existence of local minima as explained by Namgung [6]. For these reasons a designer, is required in order to validate one of the different paths found or to avoid local minima.

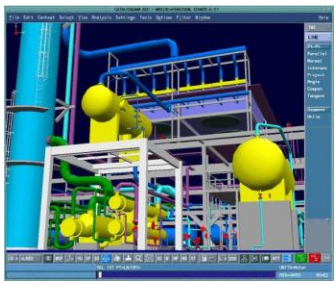 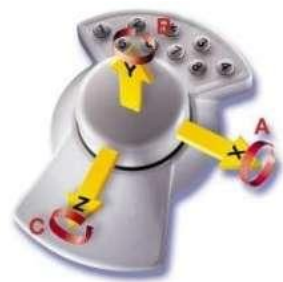

Figure 1. Manufacturing simulation.                Figure 2. SpaceMouse (LogitechTM).

The accessibility and the optimum placement of an operator to perform a task is also a matter of path planning that we propose to solve with DMU. In order to shorten time for a trajectory search, to avoid local minima and to suppress tiresome on-line manipulation, we intend to settle for a mixed approach of the above presented methodologies. Thus, we use local algorithm abilities and global view ability of a human operator, with the same approach as [7]. Among the local algorithms, we present these ones contributing to a better visibility of the task, in term of access but also in term of comfort.

## II. PATH PLANNING AND MULTI-AGENT ARCHITECTURE

The above chapter points out the local abilities of several path planners. Furthermore, human global vision can lead to a coherent partition of the path planning issue. We intend to manage simultaneously these local and global abilities by building an interaction between human and algorithms in order to have an efficient path planner [8] for a manikin or a robot with respect of ergonomic constraints or joints and mechanical limits of the robot.

### A. History

Several studies about co-operation between algorithm processes and human operators have shown the great potential of co-operation between agents. First concepts were proposed by Ferber [9]. These studies led to the creation of a "Concurrent Engineering" methodology based on network principles, interacting with cells or modules that represent skills, rules or workgroups. Such studies can be linked to work done by Arcand and Pelletier [10] for the design of a cognition based multi-agent architecture. This work presents a multi-agent architecture with human and society behavior. It uses cognitive psychology results within a co-operative human and computer system.

All these studies show the important potential of multi-agent systems (MAS). Consequently, we built a manikin "positioner", based on MAS, that combines human interactive integration and algorithms.

### B. Choice of the multi-agent architecture

Several workgroups have established rules for the definition of the agents and their interactions, even for dynamic architectures according to the environment evolution [9, 11]. From these analyses, we keep the following points for an elementary agent definition. An elementary agent:
- is able to act in a common environment,
- is driven by a set of tendencies (goal, satisfaction function, etc.),
- has its own resources,
- can see locally its environment,
- has a partial representation of the environment,

- has some skills and offers some services,
- has behavior in order to satisfy its goal, taking into account its resources and abilities, according to its environment analysis and to the information it receives.

The points above show that direct communications between agents are not considered. In fact, our architecture implies that each agent acts on its set of variables from the environment according to its goal. Our Multi Agent System (MAS) will be a black board based architecture.

*C. Path planning and MAS*

The method used in automatic path planners is schematized Figure 3a. A human global vision can lead to a coherent partition of the main trajectory as suggested in [12]. Consequently, another method is the integration of an operator to manage the evolution of the variables, taking into account his or her global perception of the environment (Figure 3b). To enhance path planning, a coupled approach using multi-agent and distributed principles as it is defined in [8] can be build; this approach manages simultaneously the two, local and global, abilities as suggested Figure 3c. The virtual site enables graphic visualization of the database for the human operator, and communicates positions of the virtual objects to external processes.

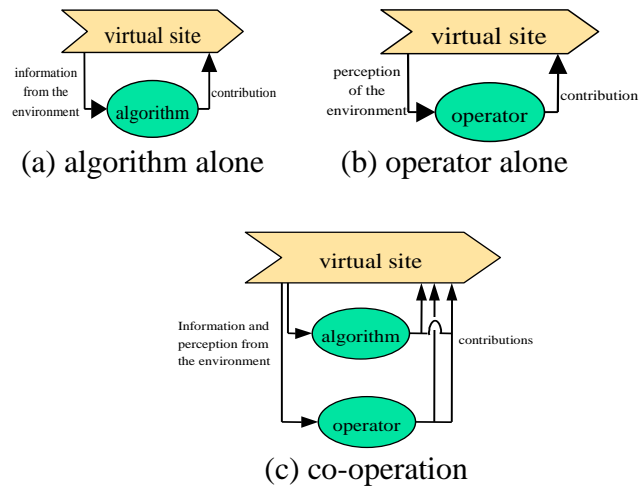

Figure 3. Co-operation principles.

As a matter of fact, this last scheme is clearly correlated with the "blackboard" based MAS architecture. This principle is described in [9, 13, 11]. A schematic presentation is presented on Figure 4. The only medium between agents is the common database of the virtual reality environment. The human operator can be considered as an elementary agent for the system, co-operating with some other elementary agents that are simple algorithms.

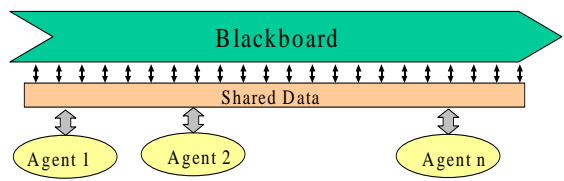

Figure 4. Blackboard principle with co-operating agents.

*D. Considered approach*

The approach we retained is the one proposed in [7] whose purpose was to validate new CAD/CAM

solutions based on a distributed approach using a virtual reality environment. This method has successfully shown its advantage by demonstrating in a realistic time the assembly task of several components with a manikin. Such problem was previously solved by using real and physical mock-ups. We kept the same architecture and developed some elementary agents for the manikin (Figure 5). In fact, each agent can be recursively divided in elementary agents.

Each agent *i* acts with a specific time sampling which is pre-defined by a specific rate of activity $\lambda i$. When acting, the agent sends a contribution, normalized by a value $\Delta_i$, to the environment and/or the manipulated object (the manikin in our study). In Figure 6, we represent the Collision agent with a rate of activity equal to 1, the Attraction agent has a rate of 3 and Operator and Manikin agents a rate of 9. This periodicity of the agent actions is a characteristic of the architecture: it expresses a priority between each of the goals of the agents. To supervise each agent activity, we use an event programming method where the main process collects agent contributions and updates the database [7]. The normalization of the actions of the agents (the values $\Delta i$) induces that the actions are relative and not absolute.

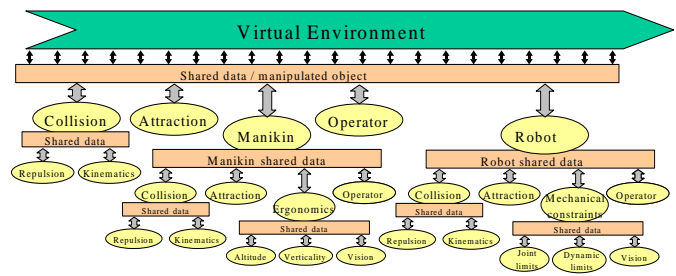

Figure 5. Co-operating agents and path planning activity [7].

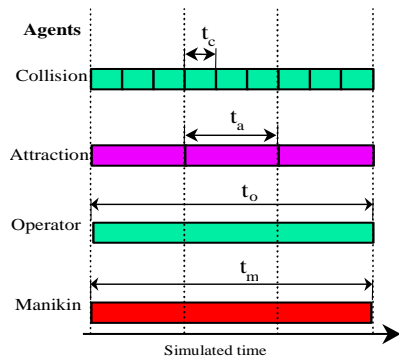

Figure 6. Time and contribution sampling

*E. Examples*

The former method is illustrated with two different examples. The first one (Figure 7) uses the MAS for testing the ability of a manikin for mounting an oxygen bottle inside an airplane cockpit through a trap. During the path planning process, the operator has acted in order to drive the oxygen bottle toward the middle of the trap. The other agents have acted in order to avoid collisions and to attract the oxygen bottle toward the final location. The real time duration is approximately 30s. The number of degrees of freedom is equal to 23.

The second example (Figure 8) is related to the automatic manipulation of a robot which base is attracted toward a wall. The joints are managed by the agents in order to avoid a collision and to solve the associated inverse kinematic model.

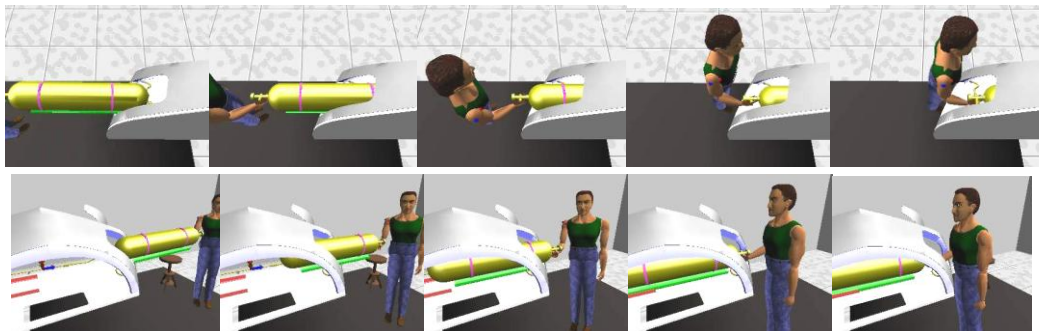

Figure 7. Trajectory path planning of a manikin using the MAS

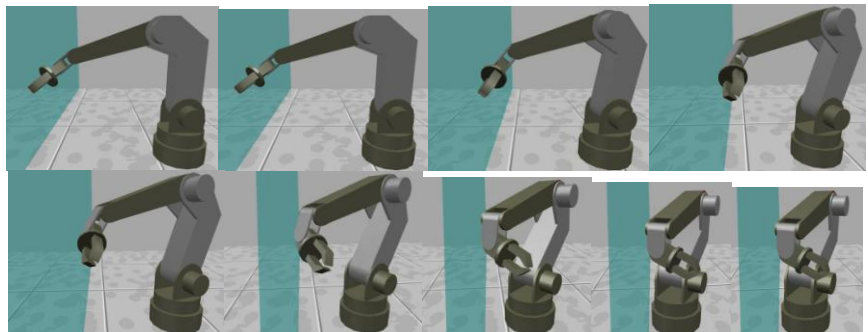

Figure 8. Trajectory path planning of a robot using the MAS

### III. VISIBILITY AND MAINTAINABILITY CHECK WITH MULTI-AGENT SYSTEM IN VIRTUAL REALITY

#### A. Introduction

For the visibility check, we have focused our attention on the trunk and the head configurations of a manikin (resp. the end-effector and the film camera orientation of a robot). The joint between the head and the trunk is characterized by three rotations $\alpha_b$, $\beta_b$ and $\theta_b$ whose range limits are defined by ergonomic constraints (Figure 9) (resp. the joint limits of the robot). These data can be found using the results of ergonomic research [14]. To solve the problem of visibility, we define a cone C whose vertex is centered between the two eyes (resp. the center of the film camera) and whose base is located in the plane orthogonal to **u**, centered on the target (Figure 10). The cone width $\varepsilon_c$ is variable.

Thus, additionally to the position and orientation variables of all parts in the cluttered environment (including the manikin itself), we consider in particular:
- Three degrees of freedom for the manikin (resp. the robot) to move it in the x-y plane: $\mathbf{x}_m = (x_m, y_m, \theta_m)^T$. It is also possible to take into account a degree of freedom $z_m$ if we want to give to the manikin (resp. the robot) the capacity to clear an obstacle.
- Three degrees of freedom for the head joint (resp. wrist joints) to manage the manikin (resp. robot) vision: $\mathbf{q}_b = (\alpha_b, \beta_b, \theta_b)^t$ with their corresponding joint constraints.

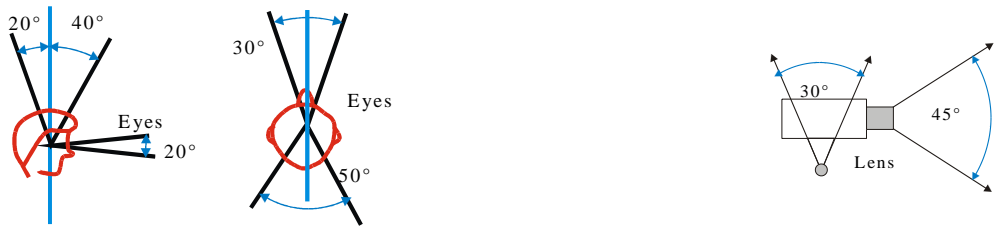

Figure 9. Example of joint limits and visibility capacity of a manikin and of a film camera.

The normalized contributions from the agents are defined with two fixed parameters: $\Delta_{pos}$ for translating moves and $\Delta_{or}$ for rotating moves.

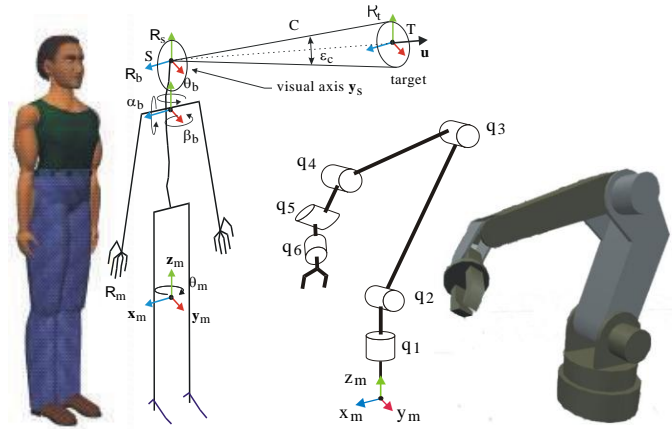

Figure 10. Manikin skeleton and robot kinematics; visibility cone and target definition.

*B. Agents ensuring visibility access and comfort for manikin, and visibility access for robot*

We present below all the elementary agents used in our system to solve the access and visibility task.
• *Attraction* agent for the manikin (resp. the robot)
The goal of the *attraction* agent is to enable the manikin (resp. the robot) to reach the target with the best trunk posture (resp. the best base placement of the robot), that is:

• To orient the projection of $y_m$ on the floor plane collinear to the projection of **u** on the same plane by rotation of $\theta_m$ (Figure 10),
• To position $x_m$ and $y_m$, coordinates of the manikin (resp. the robot) in the environment floor, as close as possible to the target position (Figure 10),
   (and for the robot.
• $q_1$ up to $q_n$ using the inverse kinematic model. This last agent acts in order to keep the robot posture, as much as possible, in the same aspect, or posture, or configuration as defined in [15].)
   This *attraction* agent only considers the target and does not take care of the environment. This agent is similar to the attraction force introduced by Khatib [4], and gives the required contributions $x_{att}$, $y_{att}$, and $\theta_{att}$ according to the attraction toward a target referenced as above. These contributions, which act on the manikin (resp. the robot) leading member position and orientation (in our case the trunk (resp. the base of the robot and its kinematics)), are normalized according to $\Delta_{pos}$ and $\Delta_{or}$.

➢ *Repulsion agent between manikin (resp. robot) and the cluttered environment*

This *repulsion* agent acts in order to avoid the collisions between the manikin (resp. the robot) and the cluttered environment, which may be static or mobile.

Several possibilities can be used in order to build a collision criterion. The intersection between two parts A and B in collision, as shown by Figure 11a, can be quantified in several ways. We can consider either the volume V of collision, or the surface Σ of collision, or the depth $D_{max}$ of collision (Figure 11b). The main drawback of these approaches comes out from the difficulty to determine these values. Moreover, 3D topological operations are not easy because many of the virtual reality softwares use polyhedral surfaces to define 3D objects. To determine $D_{move}$, the distance to avoid the collision (Figure 11b), we have to store former positions of the mobile (manikin or robot), so this quantification does not use only the database at a given instant but uses former information. This solution cannot be kept with our blackboard architecture that only provides global environment status at an instant.

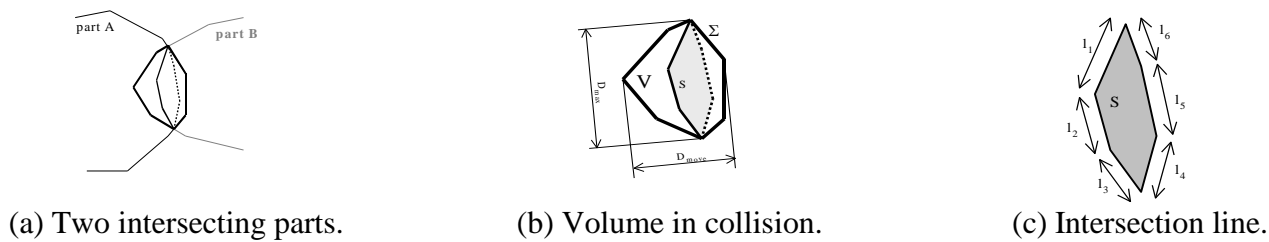

(a) Two intersecting parts.    (b) Volume in collision.    (c) Intersection line.

Figure 11. Collision criteria.

Another quantification of the collision is possible with the use of the collision line between the two parts. With this collision line, we can determine the maximum surface S or the maximum length of the collision line $l = \Sigma\, l_i$ (Figure 11c). By the end, we compute the gradient of the collision criterion according to the Cartesian environment frame using a finite difference approximation.

From the gradient vector of the collision length $\mathbf{grad}_{(x,y,\theta)}(l)$, contributions $x_{rep}$, $y_{rep}$, and $\theta_{rep}$ are computed by the *repulsion* agent. These contributions, acting on the manikin trunk position and orientation, are normalized according to $\Delta_{pos}$ and $\Delta_{or}$.

➢ *Head orientation agent*

The goal of the *head orientation* agent is to rotate the head of the manikin (resp. of the film camera) in order to observe the target. It ensures the optimum configuration that maximizes visual comfort (resp. the visibility of the target). Finding the optimum configuration consists in minimizing efforts on the joint coupling the head with the trunk and minimizing ocular efforts (resp. mechanical efforts or isotropy of the configuration). We simplify the problem by considering that the manikin has a monocular vision, defined by a cone whose principal axis, called vision axis, is along $\mathbf{y}_s$ and whose vertex is the center of manikin eyes (in that case, the manikin vision is similar to that of a film camera on a robot). If the target belongs to the vision axis, ocular efforts are considered null. Our purpose consists in orienting $\mathbf{y}_s$ such as it becomes collinear to $\mathbf{u}$ by rotation of $\alpha_b$ and $\theta_b$ (Figure 10), subject to joint limits. A joint limit average for an adult is given in Figure 9. In the case of a film camera, the corresponding values will be the optical characteristics of the film camera.

The algorithm of this agent is similar to the *attraction agent* algorithm presented there above; contributions $\alpha_{head}$ and $\theta_{head}$, after normalization, are applied to the joint coupling the head to the manikin trunk (resp. the wrist joints of the robot).

➢ Visibility *agent*

The *visibility* agent ensures that the target is visible, that is, no interference occurs between the segment ST linking the center of manikin eyes (resp. of the film camera) and the target, and the cluttered environment.

The repulsion algorithm is exactly the same as the one presented there above:

- we determine the collision line length,
- if non equal to zero, normalized contributions are determined from $x_{vis}$, $y_{vis}$, and $\theta_{vis}$ computed by the *visibility* agent according to the gradient vector of the collision length,
- contributions are applied to the manikin trunk (resp. the base of the robot).

It is to notice that some contributions may also be applied to the head orientation (resp. the film camera orientation). This is due to the fact that by turning the head, collisions between the simplified cone with the environment may also occur.

The use of a simplified cone offers the advantage of combining an ergonomic criterion with the repulsion effect. As a matter of fact, when the vision axis $\mathbf{y}_s$ is inside the cone C (Figure 10), we widen the cone, respecting a maximum limit. If not, we decrease its vertex angle, also with respect of a minimum limit that corresponds to the initial condition when starting this *visibility* agent. The maximum limit may be expressed according to the target size or/and to the type of task to perform: proximal or distant visual checking, global or specific area to control.

➢ Operator *agent on the manikin (resp. on the robot)*

One of the aims of the study is to integrate a human operator within the MAS in order to operate in real-time. The *operator* has a global view of the cluttered environment displayed by means of the virtual reality software. Her or his action must be simple and efficient. For that purpose, we use a Logitech SpaceMouse device (see /12) that allows us to manipulate a body with six degrees of freedom.

The action of the *operator* agent only considers the move of the leading object, which is in our case the manikin trunk (resp. the base of the robot). Parameters come out from position $x_{op}$ and $y_{op}$ and orientation $\theta_{op}$ returned from the SpaceMouse. These contributions are normalized, in the same way as with the *attraction* or *repulsion* agents.

IV. RESULTS AND CONCLUSIONS

This method has been tested to check the visual accessibility of specific elements under a trap of an aircraft. The digital model is presented in Figure 12 and the list of elementary agents is depicted in the master agent window in Figure 13. In this example, the *repulsion* agent for the manikin (**Repulsion**), the *visibility* agent (**Visual**) and the *head orientation* agent (**Cone**) have a specific rate of activity equal to 1, meaning that their actions have priority but it is possible for the operator to change in real-time this activity rate. Since the action of each agent is independent from the other elementary agents, it is possible to inactivate some of them (**Pause/Work** buttons). The values of $\Delta_{pos}$ and $\Delta_{or}$, which are used to normalize the agent contributions, can also be modified in real-time (**Position** and **Orientation** buttons) in order to adapt the contribution to the scale of the environment or to the task to perform.

Our experience shows that the contribution of the human operator is important in the optimization process. Indeed, if the automatic agent process fails (which can be the case when the cone used in the *visibility* agent is in collision with the environment), the human operator can:

- give to the MAS intermediate targets that will lead to a valid solution;
- move the manikin to a place where the MAS process could find a solution.

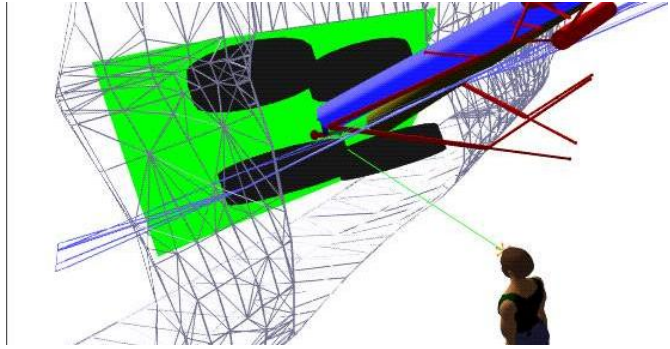

Figure 12. Digital model of a trap of an aircraft.

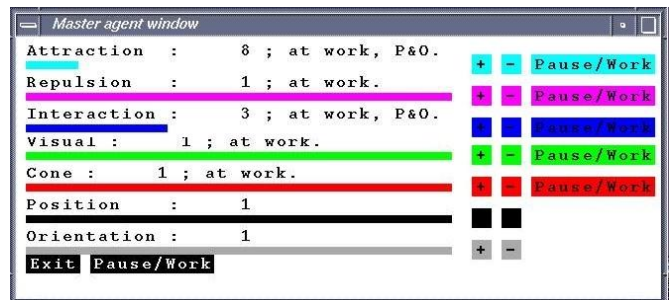

Figure 13. Master agent window.

On the other hand, the MAS allow the human operator to act more quickly and more easily with the DMU. The elementary agents guarantee a good physical and visual comfort and enable to quantify and qualify it, which would be a hard task for the human operator, even with sophisticated virtual reality devices. For instance, we can evaluate the rotations of the head and see how they are dispersed from a neutral configuration, inducing little effort. Moreover, the MAS permits a very good local collision detection and avoidance without any effort of the human operator.

The advantage of the MAS is to enable the combination of independent elementary agents to solve complex tasks. Thus, the agents participating in the visibility task can be coupled with agents enabling accessibility and maintainability as proposed by Chedmail [7]. The purpose of further works consists in a global coupling of manikin manipulations taking into account visual and ergonomic constraints and the manipulation of moving objects as robots. The result of this work is currently implemented in an industrial context with Snecma Motors [16].

To appear in Journal on Industrial, Electronics                P. Chedmail, D. Chablat, C. Le Roy[4] O. Khatib "Real time obstacle avoidance for manipulators and mobile robots," International Journal Robotics Research, pp.90-98, Vol. 5, No. 1, 1986.

[5] J. Barraquand and J. C. Latombe "Robot Motion Planning: A Distributed Representation Approach," The International Journal of Robotics research, December 1991.

[6] I. Namgung and J. Duffy "Two Dimensional Collision-Free Path Planning Using Linear Parametric Curve," Journal of Robotic Systems, pp. 659-673, Vol. 14, No. 9, 1997.

[7] P. Chedmail and C. Le Roy "A distributed approach for accessibility and maintainability check with a manikin," *Proceedings of ASME - DETC'99*, Las Vegas, September 1999.

[8] P. Chedmail, T. Damay and C. Rouchon "Integrated design and artificial reality: accessibility and path planning," ASME Engineering System Design and Analysis (ESDA'96), pp. 149-156, Vol. 8, July 1996.

[9] J. Ferber "Les systèmes multi-agents, vers une intelligence collective," *InterEditions*, 1995.

[10] J. F. Arcand and S. J. Pelletier "Cognition Based Multi-Agent Architecture," Intelligent Agent II, Agent Theories, Architectures and Languages, August 1995.

[11] N. R. Jennings, K. Sycara and M. Wooldridge, "Roadmap of agent research and development," Autonomous agents and multi-agent systems, pp. 7-38, Vol. 1, 1998.

[12] Y. K. Hwang, K. R. Cho, S. Lee, S. M. Park and S. Kang "Human Computer Cooperation in Interactive Motion Planning," ICAR'97, pp. 571-576, Monterey, July 1997.

[13] J. Ferber and M. Ghallab "Problématique des univers multi-agents intelligents," Actes des journées nationales PRC-IA, pp. 295-320, 1988.

[14] V. Riffard "Accessibilité d'un opérateur humain en environnement très contraint − Placement optimal et postures," PHD Thesis, Nantes University, France, 1995.

[15] Ph. Wenger and P. Chedmail, "Ability of a robot to travel through Cartesian free workspace in an environment with obstacles," Int. Journal of Robotics Research, vol.10,n°3, pp.214-227, Juin 1991, ISSN 0741-2223, 1988.
11/12